\documentclass[10pt,twocolumn,letterpaper]{article}

\usepackage{times}
\usepackage{epsfig}
\usepackage{graphicx}
\usepackage{amsmath}
\usepackage{amssymb}
\usepackage{inputenc, caption, subfigure, graphics, color}

\usepackage{balance}
\usepackage[normalem]{ulem}
\usepackage{multicol}
\usepackage{multirow}
\usepackage{placeins}
\usepackage{booktabs}

\usepackage{tikz}
\usepackage{filecontents}
\usepackage{pgfplots}
\usepackage{pgfplotstable}
    \pgfplotsset{compat=1.7}
\usepackage{xcolor}

\newcommand{\ARXIV}[2]{#1} 

\DeclareMathOperator{\MLP}{MLP}
\DeclareMathOperator{\PSE}{PSE}
\newcommand*{\Pa}[1]{\left(#1\right)}

\newcommand*{\upp}[2]{{#1}^{\Pa{#2}}}

\renewcommand*{\eqref}[1]{(\hyperref[#1]{\ref*{#1}})}
\newcommand*{\equaref}[1]{Equation~\hyperref[#1]{\ref*{#1}}}
\newcommand*{\figref}[1]{Figure~\hyperref[#1]{\ref*{#1}}}
\newcommand*{\tabref}[1]{Table~\hyperref[#1]{\ref*{#1}}}
\renewcommand*{\Subref}[1]{\hyperref[#1]{(\subref*{#1})}}
\newcommand*{\secref}[1]{Section~\hyperref[#1]{\ref*{#1}}}
\def\etal{\textit{et al.}~}
\def\eg{\textit{e.g.}~}

\def\ie{\textit{i.e.}~}

\def\vs{\textit{vs.}~}

\newcommand{\cS}{\mathcal{S}}

\newcommand{\tabitem}{~~\llap{\textbullet}~~}

\newcommand{\rpm}{\sbox0{$1$}\sbox2{$\scriptstyle\pm$}
 \raise\dimexpr(\ht0-\ht2)/2\relax\box2 }

\usepackage{stmaryrd}

\PassOptionsToPackage{hyphens}{url}\usepackage[breaklinks=true,bookmarks=false]{hyperref}

\begin{document}

\title{Satellite Image Time Series Classification\\ with Pixel-Set Encoders and Temporal Self-Attention}

\vspace{5cm}

\author{V. Sainte Fare Garnot$^{1}$, L. Landrieu$^{1}$, S. Giordano$^{1}$, N. Chehata$^{1,2}$  \vspace{.2cm} \\ 
$^{1}$  Univ. Gustave Eiffel, LASTIG-STRUDEL, IGN-ENSG, F-94160 Saint-Mand\'{e}, France\\
$^{2}$  EA G\&E Bordeaux INP, Universit\'e Bordeaux Montaigne, France\\
}

\maketitle

\begin{abstract}
\textit{Satellite image time series, bolstered by their growing availability, are at the forefront of an extensive effort towards automated Earth monitoring by international institutions. In particular, large-scale control of agricultural parcels is an issue of major political and economic importance. In this regard, hybrid convolutional-recurrent neural architectures have shown promising results for the automated classification of satellite image time series.We propose an alternative approach in which the convolutional layers are advantageously replaced with encoders operating on unordered sets of pixels to exploit the typically coarse resolution of publicly available satellite images. We also propose to extract temporal features using a bespoke neural architecture based on self-attention instead of recurrent networks. We demonstrate experimentally that our method not only outperforms previous state-of-the-art approaches in terms of precision, but also significantly decreases processing time and memory requirements. Lastly, we release a large open-access annotated dataset as a benchmark for future work on satellite image time series.}    
\end{abstract}

\section{Introduction}
The rising availability of high quality satellite data by both state \cite{williams2006landsat, drusch2012sentinel} and private actors \cite{maxar} opens up numerous high-impact applications for machine learning methods.
Among these, crop type classification is a major challenge for agricultural and environmental policy makers. In the European Union (EU), yearly crop maps are needed to grant the Common Agricultural Policy subsidies, an endowment of over $50$ billion euros each year \cite{europa1}. Currently, European farmers declare the cultivated species manually on a yearly basis.
The EU's Joint Research Center has thus called for the development of efficient tools to achieve automated monitoring \cite{europa2}. This push to automation is motivated in part by the launch of the \text{Sentinel-$2$} satellite---which became fully operational in mid-$2017$---by the European Space Agency \cite{drusch2012sentinel}, and whose settings are particularly valuable for crop classification. Indeed, its high spectral resolution (13 bands) and short revisit time of $5$ days are well-suited to analysing crop \emph{phenology}, \ie the cyclical evolution of vegetation \cite{vrieling2018vegetation}. Additionally, the farmers' yearly manual declarations provide a considerable amount of annotated data ($10$ million parcels labelled each year in France alone) to train learning algorithms. Such models would have a wide array of applications beyond crop monitoring, for both public and private entities. 

Practitioners mainly rely on traditional methods such as Random Forest (RF) and Support Vector Machine (SVM), which operate on handcrafted features for automated crop classification \cite{inglada2015assessment,zheng2015support}. Recently, the gradual adoption of deep learning methods such as Convolutional Neural Networks (CNN) and Recurrent Neural Networks (RNN) for learning spatial and temporal attributes has brought significant improvements in classification performance.
More specifically, hybrid neural architectures combining convolutions and recurrent units in a single architecture are the current state-of-the-art for  crop type classification \cite{russwurm2018convolutional,garnot2019time}.

In this paper, we argue that such hybrid recurrent-convolutional architectures fail to adapt to some key characteristics of the problem under consideration.
\vspace{-.29cm}
\paragraph{Spatial Encoding of Parcels:} Sensors typically used for crop classification, such as the \text{Sentinel-$2$} satellites, have a coarser spatial resolution ($10$m per pixel) than the typical agricultural textural information such as furrows or tree rows. However, CNNs rely heavily on texture to extract spatial features \cite{CNNtexture}.
Given this limitation, we propose to view medium-resolution images of agricultural parcels as unordered sets of pixels. Indeed, recent advances in 3D point cloud processing have spurred the development of powerful encoders for data comprised of sets of unordered elements \cite{qi2017pointnet, zaheer2017deep}.
\begin{figure*}
\begin{center}
    \includegraphics[width=.8\linewidth, trim = 0cm 17.9cm 1cm  5.1cm , clip]{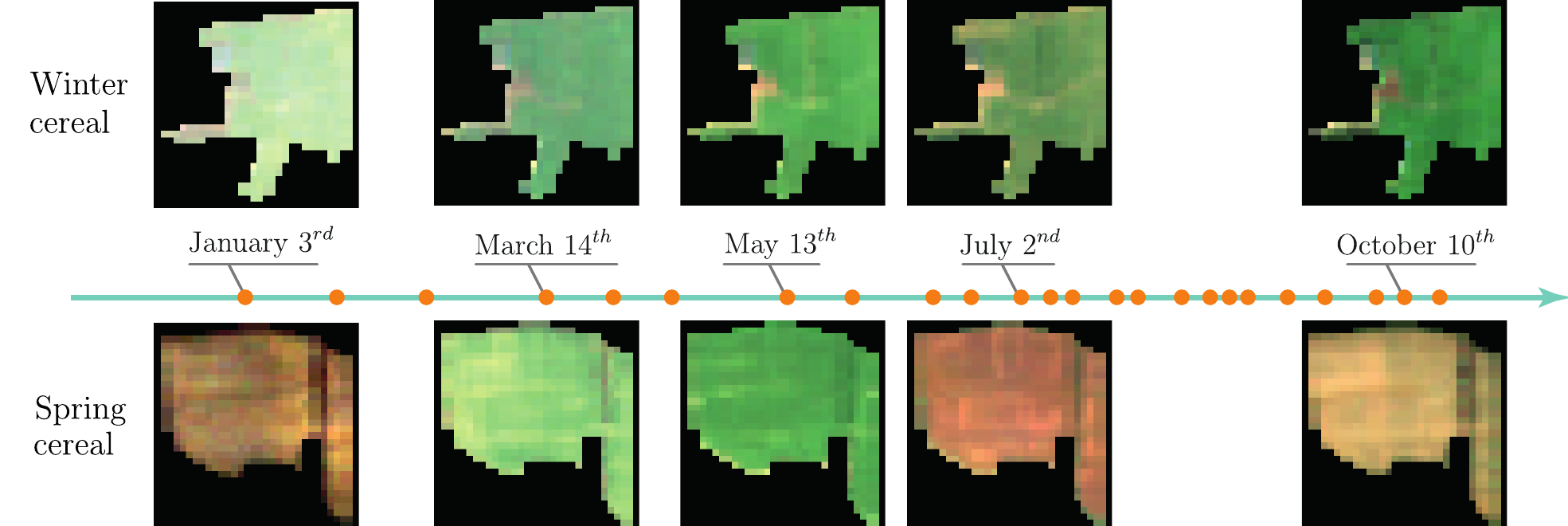}
\end{center}
   \caption{Example of \text{Sentinel-$2$}  time series (shown: RGB bands, $10$m per pixel) for two parcels of the \textit{Winter cereal} and \textit{Spring cereal} classes. The dots on the horizontal axis represent  the unevenly distributed acquisition dates over the period of interest.
   Note the importance of the temporal evolution of the parcels to discriminate between the classes.
   }
    \label{fig:ITS}
\end{figure*}
We show in this paper that set-based encoders can successfully extract learned statistics of the distribution of spectra across the spatial extent of the parcels. Furthermore, we show that this approach handles the highly-variable size of parcels in a more efficient way than CNNs.
\vspace{-.29cm}
\paragraph{Temporal Encoding of Satellite Time Series:} Earlier work in crop classification has shown the importance of the temporal dimension when classifying crop types \cite{garnot2019time}. While RNNs have been widely used to analyse temporal sequences, recent work in Natural Language Processing (NLP) has introduced a promising new approach based on attention mechanisms \cite{vaswani2017attention}. The improved parallelism brought by this approach is particularly valuable for automated crop monitoring, as its typical scale spans entire continents: one year of Sentinel-2 observations amounts to $25$Tb of data for agricultural areas in the EU.
Therefore, we propose an adapted attention-based approach for the classification of time series.~\\

\noindent
The key contributions of this paper are as follows:
\begin{itemize}
    \item Inspired by Qi \etal \cite{qi2017pointnet}, we introduce the pixel-set encoder
    as an efficient alternative to convolutional neural networks for medium-resolution satellite images.
    \item We adapted the work of Vaswani \etal \cite{vaswani2017attention} to an end-to-end, sequence-to-embedding setting for time series.
    \item We establish a new state-of-the-art for the task of large-scale agricultural parcel classification. Moreover, our method not only improves the classification precision by a significant margin, but simultaneously boasts a acceleration of over $4$ times and a memory imprint reduced by over $70\%$ compared to the best-performing approaches in the literature.
    \item We release the first open-access dataset of \text{Sentinel-$2$} images for crop classification with ground truth labels.
\end{itemize}
\section{Related Work}
\label{sec:rw}
The problem of satellite image time series classification can be addressed at pixel level or object level. Pixel-based approaches do not require \emph{a priori} knowledge of the borders of parcels, but cannot leverage the spatial homogeneity of class labels within the object's extent. Conversely, in the case of crop classification, object-based approaches can leverage the parcels' shape to extract helpful spatial information for achieving better classifications \cite{devadas2012support}.
\vspace{-.29cm}
\paragraph{Traditional Machine Learning:} Until recently, the common approach for crop classification has been to use traditional discriminative models with handcrafted features \cite{vuolo2018much,inglada2015assessment,wardlow2008large}. For instance, the Normalized Difference Vegetation Index (NDVI) combining the red and near-infrared spectral bands has been widely used as it relates to crop photosynthetic activity \cite{tucker1979red}. Certain work also includes phenological features derived from the study of the NDVI as well as external meteorological information \cite{zhong2014efficient}. Although robust and easily interpretable, such handcrafted indices do not compare favorably to end-to-end learned features.

In such work, the prevalent approach to represent temporal evolution is to concatenate each date's spatial and spectral features. This is not well-suited to application over large geographical areas, in which the  acquisition dates vary depending on the satellite orbit, and in which cloud cover and meteorological condition can be heterogeneous, resulting in sequences of variable length and temporal sampling. Consequently, other work oriented their efforts towards a better modeling of time using Hidden Markov Models \cite{siachalou2015hidden}, Conditional Random Fields \cite{bailly2018crop}, or Dynamic Time Warping \cite{belgiu2018sentinel}. 
\vspace{-.29cm}
\paragraph{Convolutional and Recurrent Approaches:} More recently, the successful advances in the deep learning literature have provided efficient tools for both spatial and temporal feature extraction. Although some work only uses these tools as feature extractors \cite{nijhawan2017deep}, or  combine them with feature engineering \cite{zhang2018mapping}, most current work follows the deep learning paradigm of end-to-end trainable architectures. More specifically, Kussul \etal \cite{kussul2015parcel} proposed to use a Multi Layer Perceptron (MLP) on raw observation data instead of traditional RF of SVM. Further work sets out to leverage the spatial and temporal structures of time series of satellite images. CNNs \cite{lecun1995convolutional} appeared to be a natural choice to address the spatial dimensions of the data \cite{kussul2017deep,russwurm2018multi}. Similarly, Long-Short Term Memory (LSTM) networks \cite{hochreiter1997long} were successfully applied to model the temporal dimension of the data \cite{russwurm2017temporal,ndikumana2018deep}, outperforming RF and SVM \cite{ienco2017land}.

Furthermore, Ru{\ss}wurm \etal \cite{russwurm2018convolutional} first proposed to use hybrid recurrent convolutional approach by applying the ConvLSTM architecture \cite{xingjian2015convolutional} to parcel classification. This work yielded state-of-the-art results and also showed that ConvLSTMs are able to learn to detect and ignore cloud obstruction.  A similar approach was successfully used for automated change detection from Sentinel-2 data as well \cite{papadomanolaki2019detecting}.
Finally, Garnot \etal showed in \cite{garnot2019time} that higher classification performance can be obtained by implementing such a hybrid model but with two dedicated modules for spatial and temporal feature extraction respectively: the series of images is first embedded by a shared CNN and the resulting embeddings sequence is fed to a Gate Recurrent Unit (GRU) \cite{chung2014GRU}. The use of a GRU is motivated by the smaller number of parameters required to achieve similar performance as LSTM, as corroborated in \cite{russwurm2018multi}. Additionally, Garnot \etal show that the relatively low spatial resolution of multi-temporal satellite images may question the relevance of CNNs since handcrafted descriptors of spectral distribution performed nearly as well as trainable spatial encoders when used in combination to the recurrent units. This is one of the issues we propose to address in the present study. 
\vspace{-.29cm}
\paragraph{Attention-Based Approach:}
Following the adoption of self-attention in the NLP literature as an efficient  alternative to RNNs, Ru{\ss}wurm \etal proposed in \cite{russwurm2019self} to apply the Transformer architecture \cite{vaswani2017attention}---a self-attention based network---to pixel-based classification. Their extensive experiments show that the Transformer yields classification performance that is on par with RNN-based models and present the same robustness to cloud-obstructed observations. Likewise, we propose to extend self-attention mechanisms to end-to-end sequence-to-embedding learning on images for object-level classification.
\vspace{-.29cm}
\paragraph{Purely Convolutional Approach:}
Multiple papers propose to address the temporal dimension with convolutions.
Ji \etal present in \cite{ji20183d} a spatio-temporal 3D-CNN for parcel-based classification, and spectro-temporal convolutions are found to outperform LSTMs for pixel-based segmentation on temporal profiles in \cite{pelletier2019temporal}, and outperform an MLP in \cite{kussul2017deep}. Similar results are found in  \cite{zhong2019deep}, where temporal convolutions yield better results than an LSTM network for classification based on NDVI temporal profiles. 
In addition, temporal convolutions have significantly lower processing times than RNNs. Yet, the ability to account for long-term dependencies requires deeper architectures. Furthermore, the fixed architecture of temporal CNN prevents the same network from being used on sequences of different lengths or with different acquisition dates.

Lastly, 2D and 3D convolutions have been extensively used  in video analysis for object segmentation \cite{caelles2017one,shin2017pixel} or action recognition \cite{carreira2017quo, feichtenhofer2016convolutional}. However, specificities of satellite time series such as their different time-scale and resolution prevents the direct application of such networks.

\section{Methods}
In this section, we present the different components of our proposed architecture for encoding time series of medium-resolution multi-spectral images. We denote the observations of a given parcel by a spatio-spectro-temporal tensor $[\upp{x}{0}, \cdots, \upp{x}{T}]_{t=1}^T$ of size $T\times C \times H \times W$, 
with $T$ the number of temporal observations, $C$ the number of spectral channels, and $H$ and $W$ the dimension in pixels of a tight bounding box containing the spatial extent of the parcel. All values are set to $0$ outside the parcel's borders, as shown in \figref{fig:ITS}.

\begin{figure*}[h!]
    \includegraphics[width=\textwidth, trim = 0cm 5.5cm 0cm 5cm ]{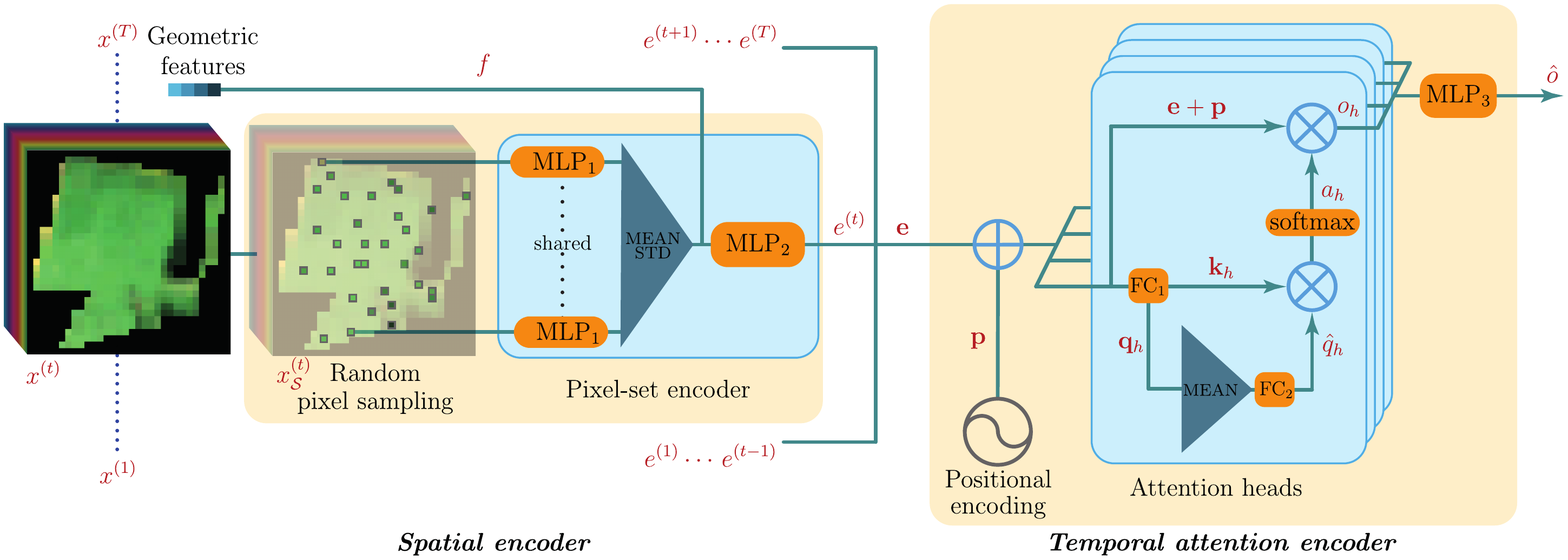}
    \caption{Schematic view of our spatio-temporal encoder. Variables in bold are tensors concatenated along the temporal dimension, \eg $\mathbf{e}=[\upp{e}{0}, \cdots, \upp{e}{T}]$.}
    \label{fig:model}
\end{figure*}
\subsection{Spatial Encoder}
In recent years, CNNs have become the established approach to extract spatial features from images.
However, our analysis suggests that convolutions may not be well-suited for the analysis of medium-resolution satellite images of agricultural parcels.
Indeed, as mentioned above, the typical spatial resolution of satellites with high revisit frequency struggles to capture textural information. Second, efficiently training CNNs requires organizing the data into batches of images of identical dimensions. The irregular size of the parcels makes this process very memory intensive. Indeed, to limit textural information loss for large parcels, this amounts to oversampling most smaller parcels several times over.

To circumvent both these issues, we propose an alternative architecture called \emph{Pixel-Set Encoder} (PSE) and inspired by the point-set encoder PointNet \cite{qi2017pointnet} and the Deep-Set architecture \cite{zaheer2017deep} commonly used for 3D point cloud processing. 
The motivation behind this design is that, instead of textural information, the network computes learned statistical descriptors of the spectral distribution of the parcel's observations.

The network proceeds as follows to embed an input observation $\upp{x}{t}$:
\begin{itemize}
    \item[i)] A set $\cS\subset [1,\cdots,N]$ of $S$ pixels
    is randomly drawn from the $N$ pixels within the parcel, as described in \equaref{eq:pse:sample}. When the total number of pixels in the image is smaller than $S$, an arbitrary pixel is repeated to match this fixed size. 
    The same set $\cS$ is used for sampling all $T$ acquisitions of a given parcel. 
    \item[ii)] Each sampled pixel $s$ is processed by a shared multi-layer perceptron $\MLP_{1}$, as seen in \equaref{eq:pse:mlp1}, composed of a succession of fully-connected layers, batchnorms \cite{ioffe2015batch}, and Rectified Linear Units \cite{nair2010rectified}.
    \item[iii)] The resulting set of values is pooled along the pixel axis---of dimension $S$---to obtain a vector capturing the statistics of the whole parcel and which is invariant by permutation of the pixels' indices. We concatenate to this learned feature a vector of pre-computed geometric features $f$:  perimeter, pixel count $N$, cover ratio ($N$ divided by the number of pixels in the bounding box) and the ratio between perimeter and surface of the parcel.
    \item[iv)] This vector is processed by another perceptron $\MLP_{2}$, as shown in \equaref{eq:pse:mlp2}, to yield $\upp{e}{t}$
    the parcel's spatio-spectral embedding at time $t$. 
\end{itemize}
The PSE architecture is represented in \figref{fig:model}, and can be summarized by the following equations:
\begin{eqnarray}
\label{eq:pse:sample}
    \cS &\!=\!&\text{sample}\Pa{S,N}\\
    \label{eq:pse:mlp1}
    \upp{\hat{e}}{t}_s &\!=\!& \MLP_1
    \Pa{\upp{{x}}{t}_s}\;,\;\forall s \in \cS \\
    \label{eq:pse:mlp2} 
    \upp{e}{t} &\!=\!& \MLP_{2}
    \Pa{
        \left[
            \text{pooling}
            \Pa{\{\upp{\hat{e}}{t}_s\}_{s\in\cS}}
              ,f
        \right]
    }~.
\end{eqnarray}
Among possible pooling operations, we had the best results for the concatenation of the mean and the standard deviation across the sampled pixel dimension $S$. For parcels smaller than $S$, repeated pixels should be removed before pooling to obtain unbiased estimates. 

Although only a limited amount of information per parcel is used by this encoder, the sampling being different at each training step ensures the learning of robust embeddings exploiting all the available information.
%
\subsection{Temporal Attention Encoder}
RNNs have proven efficient for encoding sequential information \cite{lipton2015critical}. However, since RNNs process the elements of the sequence successively, they prevent parallelization and incur long training times. In \cite{vaswani2017attention}, Vaswani \etal introduce the Transformer architecture, an attention-based network achieving equal or better performance than RNNs on text translation tasks, while being completely parallelizable and thus faster. We propose to adapt their 
ideas to the the encoding of satellite image time series.
\vspace{-.29cm}
\paragraph{Transformer Network:}
In the original Transformer model a \emph{query-key-value} triplet $\Pa{ \upp{q}{t},\upp{k}{t},\upp{v}{t}}$ is computed simultaneously for each element of the input sequence by three fully-connected layers. 
For a given element of a sequence, the key $\upp{k}{t}$ conveys information about the nature of its content, while the value $\upp{v}{t}$ encodes the content itself. The output of a given element is defined as the sum of the values of previous elements weighted by an attention mask. This mask is defined as the compatibility (dot product) of the keys of the previous elements with the query $\upp{q}{t}$, re-scaled through a modified softmax layer. In other words, each element indicates which kind of information it needs through its query, and what sort of information it contains through its key.

Since the computation of the triplets $\Pa{\upp{q}{t}, \upp{k}{t}, \upp{v}{t}}$ and their multiplications can be performed in parallel, the Transformer takes full advantage of modern GPU architecture and boasts a significant speed increase compared to recurrent architectures.
This procedure can be computed several times in parallel with different set of independent parameters, or \emph{heads}. This approach, called \emph{multi-head attention}, allows for the specialization of different set of query-key compatibility.

\vspace{-.29cm}
\paragraph{Positional Encoding:} In their paper on text translation, Vaswani \etal add order information to elements of the input sequence by adding a positional encoding tensor to each element.  \equaref{eq:pe} describes this positional encoding of the observation $t$, with $d_{e}$ the dimension of the input, and $i$ the coordinates of the positional encoding. Since our considered sequences are typically shorter than the ones considered in NLP, we chose $\tau= 1\,000$---instead of $10\,000$. Additionally, $\text{day}(t)$ is the number of days since the first observation for observation $t$ instead of its index. This helps to account for inconsistent temporal sampling (see \figref{fig:ITS}).
\begin{align}
[\upp{p}{t}]_{i=1}^{d_e} = \text{sin}\Pa{{\text{day}(t)}\backslash{ \tau^{\frac{2i}{d_{e}}}}          
+\frac\pi2 \text{mod}(i,2)}\label{eq:pe}
\end{align}
\vspace{-.4cm}
\paragraph{End-to-End Encoding:}
The original Transformer network takes pretrained word embeddings as inputs. In our setting however,
the parameters of the network producing the inputs  are learnt simultaneously to the attention parameters.
Therefore, we propose that  each head only computes key-query pairs from the spatial embeddings \eqref{eq:trans:fc} since these embeddings can directly serve as values: $v^{(t)}=e^{(t)}+p^{(t)}$. This removes needless computations, and avoids a potential information bottleneck when computing the values.
\vspace{-.29cm}
\paragraph{Sequence-to-Embedding Attention:} While the original Transformer produces an output for each element of a sequence, our goal is to encode an entire time series into a single embedding. 
Consequently, we only retain the \emph{encoder} part of the Transformer and define a single \emph{master query} $\hat{q}_h$ for each head $h$.  Such a query, in combination with the keys of the elements of the sequence, determines which dates contain the most useful information. A first approach would be to select the query of a given date, such as the last one. However, the selected  element of the sequence may not contain enough information to produce a meaningful query. Instead, we propose to construct the master query as a temporal average of the queries of all dates and processed by a single fully-connected layer \eqref{eq:trans:master}. As shown in \equaref{eq:trans:attention}, this query is then multiplied with the keys of all elements of the sequence to determine a single attention mask $\upp{a}{h}\in[0,1]^T$, in turn weighting the input sequence of embeddings \eqref{eq:trans:output}.

\vspace{-.29cm}
\paragraph{Multi-Head Self-Attention:} 
We concatenate the output $o_{h}$ of each head $h$ for the $H$ different heads and process the resulting tensor with $\MLP_3$, to obtain the final output $\hat{o}$ of the Temporal Attention Encoder (TAE), as shown in \equaref{eq:trans:multihead}. Note that unlike the Transformer network, we directly use $\hat{o}$ as the spatio-temporal embedding  instead of using residual connections.

\vspace{-.29cm}
\paragraph{Temporal Attention Encoder}
For each head $h$, we denote by $\upp{\text{FC}}{h}_{1}$ the fully-connected layer generating the key-query pairs, $\upp{\text{FC}}{h}_{2}$ the fully-connected layer yielding the master query, and $d_k$ the shared dimensions of  keys and queries. Our attention mechanism can be summarized by the following equations for all $t \in [1,\dots,T]$ and $h \in [1, \cdots, H]$:
\begin{align}\label{eq:trans:fc}
&\upp{k}{t}_{h},\upp{q}{t}_{h} = 
\upp{\text{FC}}{h}_{1}\Pa{\upp{e}{t} + \upp{p}{t}}\\\label{eq:trans:master}
&\hat{q}_{h} =  \upp{\text{FC}}{h}_{2}\left(\text{mean}\Pa{\{\upp{q}{t}_{h}\}_{t=1}^T}\right) \\\label{eq:trans:attention}
&a_{h}   = \text{softmax}\Pa{\frac1{\sqrt{d_{k}}}\left[\hat{q}_{h} \cdot \upp{k}{t}_{h}\right]_{t=1}^T}
\\\label{eq:trans:output}
&o_{h} =  \sum_{t = 1}^{T} a_h[t] \Pa{\upp{e}{t} + \upp{p}{t}} \\\label{eq:trans:multihead}
&\hat{o} = \MLP_{3}\Pa{[o_{1}, \cdots, o_{H}]}~.
\end{align}
\subsection{Spatio-Temporal Classifier}
Our spatio-temporal classifier architecture combines the two components presented in the previous sections:
all input images of the time series are embedded in parallel by a shared $\PSE$, and the resulting sequence of embeddings is processed by the temporal encoder, as illustrated in \figref{fig:model}. Finally, the resulting embedding is processed by an MLP decoder  $\MLP_4$ to produce class logits $y$:
\begin{align}
    y=\MLP_4 \Pa{\hat{o}}~.
\end{align}
%
%
\subsection{Implementation details}
%
All the architectures presented here are implemented in PyTorch, and released on GitHub upon publication.\footnote{\url{github.com/VSainteuf/psetae}} We trained all models on a machine with a single GPU (Nvidia 1080Ti)  and an 8-core Intel i7 CPU for data loading from an SSD hard drive. We chose the hyperparameters of each architecture presented in the numerical experiments such that they all have approximately $150$k trainable parameters. 
The exact configuration of our network is displayed in \tabref{tab:hyperp_agrinet}.
We use the Adam optimizer \cite{kingma2014adam} with its default values ($lr={10}^{-3}, \quad \beta=(0.9, 0.999$)) and a batch size of $128$ parcels. We train the models with focal loss \cite{lin2017focal} ($\gamma = 1$) and implement a $5$-fold cross-validation scheme: for each fold the dataset is split into train, validation, and test set with a 
3:1:1 ratio. The networks are trained for $100$ epochs, which is sufficient for all models to achieve convergence. We use the validation step to select the best-performing epoch, and evaluate it on the test set. For augmentation purpose, we add a random Gaussian noise to $\upp{x}{t}$ with standard deviation ${10}^{-2}$ and clipped to $5.{10}^{-2}$ on the values of the pixels, normalized channel-wise and for each date individually.
\section{Numerical Experiments}
\begin{table}[]
\begin{tabular}{llr}
 \multirow{2}{*}{Modules}& \multirow{2}{*}{Hyperparameters}                                                 &  Number of \\
 &    &      parameters            \\ \midrule
\textbf{PSE} &   &        19\,936       \\\toprule
S & 64 & \\
$\text{MLP}_{1} $ & $10 \rightarrow32\rightarrow64$  & \\ 
$\text{MLP}_{2} $ & $ 68 \rightarrow128$  & \\ \midrule
\textbf{TAE}       &      &                116\,480  \\ \toprule
$d_{e} $, $d_{k} $,  $H$  &      $128$, $32$, $4$ &                  \\ 
$\text{FC}_{1} $ &    $128 \rightarrow (32 \times 2)$  &                  \\ 
$\text{FC}_{2} $ &    $32 \rightarrow 32$  &                  \\ 

$\text{MLP}_{3} $              &    $512 \rightarrow 128\rightarrow128$  &                  \\ 
\midrule
\textbf{Decoder}           &      &      11\,180            \\ \toprule
$\text{MLP}_{4} $ & $128\rightarrow64\rightarrow32 \rightarrow 20$  &  \\ \toprule
\textbf{Total} & & 147\,604
\end{tabular}
\caption{Configuration of our model chosen for the numerical experiments. The dimension of each successive feature space is given for MLPs and fully connected layers. We show the corresponding number of trainable parameters on the last column.  }
\label{tab:hyperp_agrinet}
\end{table}
\subsection{Dataset}
We evaluate our models using \text{Sentinel-$2$} multi-spectral image sequences in top-of-canopy reflectance. We leave out the atmospheric bands (bands $1$, $9$, and $10$), keeping $C=10$ spectral bands. The six $20$m-resolution bands are resampled to the maximum spatial resolution of 10m.

The area of interest (AOI) corresponds to a single tile of the \text{Sentinel-$2$} tiling grid (T31TFM) in southern France. This tile provides a challenging use case with a high diversity of crop type and different terrain conditions. The AOI spans a surface of $12 \, 100$$\:$km$^2$   and contains $191 \, 703$ individual parcels, all observed on $24$ dates from January to October 2017. The values of cloudy pixels are linearly interpolated from the first previous and next available pixel using Orfeo Toolbox \cite{christophe2008orfeo}. 

\begin{figure}[h]
\begin{tikzpicture}
 \pgfplotstableread
 [col sep=comma]
 {Class_rep4320.csv}{\loadedtable}
    \begin{semilogyaxis}[
        width=\axisdefaultwidth,
        height=.65*\axisdefaultheight,
        ymin=1e1,
        ymax=3*1e5,
        ybar=1pt,
        xtick=data,
        xticklabels from table={\loadedtable}{ENG},
        xticklabel style={rotate=45, anchor=east, font=\scriptsize},
        log origin=infty,
        bar width=0                 .5,
        enlarge x limits={abs=0.6},
    ]
       \addplot table [
                x expr=\coordindex,
                y =count,
                col sep=comma
            ] {\loadedtable};

        \addlegendentry{Class count}
    \end{semilogyaxis}
    \end{tikzpicture} 
    \caption{Class repartition in the AOI.}
    \label{fig:class_rep}
\end{figure}
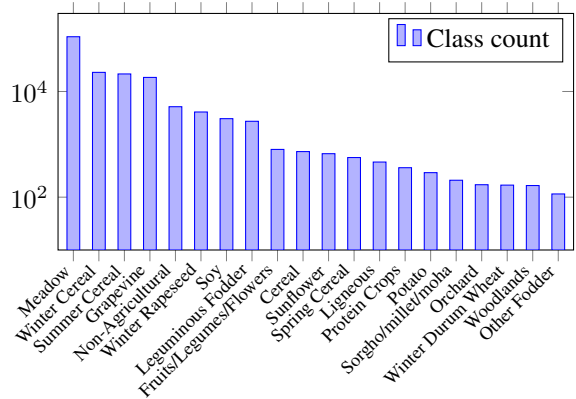

We retrieve the geo-referenced polygon and class label of each parcel from the French Land Parcel Identification System records.\footnote{\url{http://professionnels.ign.fr/rpg}} We crop the satellite images using this polygon to constitute the image time series.
\vspace{-.29cm}
\paragraph{Data Preparation:} In order to evaluate both ours and convolution-based methods, we organize the parcels into two different formats: patches and pixel sets. 

In the \emph{patch} format, we resize each parcel into a tensor of size $ T \times C \times 32\times 32$ by interpolating each spectral channel and temporal acquisition independently into patches of fixed size $32 \times 32$. We use nearest neighbor interpolation, and both the horizontal and vertical axes are rescaled so that the overall shape of the parcel may be altered. We use zero-padding  outside the extent of the parcel (see Figure \ref{fig:ITS}). 
This same size of $32$ pixels was used in \cite{garnot2019time}, while a larger $48\times48$ patch size was used in \cite{russwurm2018convolutional}, albeit for a pixel-wise classification task.

For the \emph{pixel-set} format, the pixels of each parcels are stored in arbitrary order into a tensor of size $T \times C \times N$, with $N$ the total number of pixels in a given parcel. Note that this format neither lose nor create information, regardless of parcel size. Hence, this setup saves up to $70\%$ disk space compared to the patch format ($28.6$Gb \vs$98.1$Gb). Note that the geometric features $f$ must be computed and saved before preparing the dataset, as all spatial structure is henceforth lost.

The classification labels are defined with respect to a $20$ class nomenclature designed by the subsidy allocation authority of France. We show the class break-down on the AOI in  \figref{fig:class_rep}. The dataset is highly imbalanced as is often the case in such real word applications and this motivated the use of the focal loss to train our models.

Both datasets will be released upon publication.\footnote{\url{github.com/VSainteuf/psetae}} To the best of our knowledge, no benchmark dataset currently exists for object-based agricultural parcel classification. Our datasets are a first step towards more reproducible and comparable methodological work in this field.
\begin{table*}[h]
\begin{center}
\begin{tabular}{l|ccccc}
& \multirow{2}{*}{OA}   & \multirow{2}{*}{mIoU}  & Training   & Inference & Disk Size \\
& &  &  (s/epoch) &  (s/dataset) & Gb  \\ \toprule
PSE+TAE (ours)                     &\textbf{94.2} \footnotesize{\rpm 0.1}  &  \textbf{50.9}  \footnotesize{\rpm 0.8}   & {158} & \bf 149 & \bf 28.6 / 12.3$^1$\\ 
CNN+GRU  \cite{garnot2019time}                  & 93.8 \footnotesize{\rpm 0.3}    &   48.1  \footnotesize{\rpm 0.6}   &    656 & 633 & 98.1\\
CNN+TempCNN \cite{pelletier2019temporal}        & 93.3\footnotesize{\rpm 0.2}    &   47.5\footnotesize{\rpm 1.0}    &      635 & 608 & 98.1\\ 
Transformer \cite{russwurm2019self}                    &  93.0 \footnotesize{\rpm 0.2 }  &  46.3  \footnotesize{\rpm  0.9} & 13  & 420 + 4$^3$ &\bf 28.6 / 0.22$^4$ \\
ConvLSTM  \cite{russwurm2018convolutional}      & 92.5\footnotesize{\rpm 0.5}    &   42.1\footnotesize{\rpm 1.2}   &  1\,283 & 666 & 98.1 \\
Random Forest \cite{bailly2018crop}        & 91.6\footnotesize{\rpm 1.7}   & 32.5  \footnotesize{\rpm 1.4} & \bf 293$^2$ & {420 + 4}$^3$ & \bf 28.6 / 0.44 $^4$
\end{tabular}
\end{center}
\vspace{-.29cm}
\caption{Classification metrics and time benchmark of the different architectures. The inter-fold standard deviation of the OA and mIoU is given in smaller font. Additionally, the total time for one epoch of training, and for inference on the complete dataset are given on the third and fourth columns. \small{ $^1$ disk space required for training and pure inference, $^2$ time for the entire training step, $^3$ preprocessing and inference time, $^4$ dataset before and after preprocessing.} }
\label{tab:perf_classif}
\centering
\end{table*}
\vspace{-.29cm}
\subsection{Comparison with State-of-the-Art}
\paragraph{Competing Methods:} We compare our approach to recent algorithms operating on similar dataset, which we have re-implemented. The different hyperparameters chosen for each model are shown in the appendix. All share the same decoding layer configuration $\MLP_{4}$.
\begin{itemize}
    \item[]\textbf{CNN+GRU} In \cite{garnot2019time}, Garnot \etal propose a similar approach to ours, but with CNNs instead of PSE, and GRUs instead of our proposed temporal encoder. The last hidden state of the recurrent unit is used as input to $\MLP_{4}$ for classification. 
    \item[]\textbf{CNN+TempCNN} In \cite{pelletier2019temporal}, Pelletier \etal propose to use one-dimensional temporal convolution to address the sequential nature of the observations. While their approach is applied on a per-pixel classification task and therefore not comparable, we have implemented a variation of CNN+GRU in which the GRUs are replaced with one-dimensional convolutions as the closest translation of their ideas. 
    \item[]\textbf{Transformer} In \cite{russwurm2019self}, Ru{\ss}wurm \etal perform object-based classification with the encoder part of the Transformer network. They do not use a spatial encoder and compute average values of the different spectral bands over each parcel. Furthermore they  produce a single embedding for the whole sequence with a global maximum pooling through the temporal dimension of the output sequence. We re-implemented the same pipeline and simply modified the hyperparameters to match the $150$k parameters constraint.
    \item[]\textbf{ConvLSTM} In \cite{russwurm2018convolutional}, Ru{\ss}wurm \etal process the time series of \textit{patch} images with a ConvLSTM network \cite{xingjian2015convolutional} for pixel-based classification. We adapt the architecture to the parcel-based setting by using the spatially-averaged last hidden state of the ConvLSTM cell to be processed by $\MLP_{4}$.
    \item[]\textbf{Random Forest} Lastly, we use a Random Forest classifier  with $100$ trees as a non-deep learning baseline. The classifier operates on handcrafted features comprised of the mean and standard deviation of each band within the parcel, and concatenated along the temporal axis, as described by \cite{bailly2018crop}. 
\end{itemize}
We present the results of our experiments in \tabref{tab:perf_classif}.
Our proposed architecture outperforms the other deep learning models in Overall Accuracy (OA) by $0.4$ points, and mean per-class Intersect over Union (mIoU) by $3$ to $9$ points. It also provides a four-fold speed up over convolution-based methods, and a decrease in disk usage of over $70\%$ for training, and close to $90\%$ when considering the inference task alone (\ie when only $S$ pixels per parcels are kept). This speed-up is due both to improved loading time as the pixel set dataset is smaller, but also inference and backpropagation time, as detailed in Table~2 of the appendix. While the temporal convolutions of TempCNN are faster to train, they yield worse performance and suffer from the limitations discussed in section \ref{sec:rw}. The Transformer method, which processes pre-computed parcel means, is also faster to train, but only achieves a $46.3$ mIoU score.

Beyond its poor precision, the RF classifier has a significant speed and memory advantage. This can explain its persisting popularity among practitioners.
However, our approach bridges in part this performance gap and provides much higher classification rates, making it a compelling strategy for large-scale object-based crop type mapping.
%
\subsection{Ablation Studies}
In order to independently assess the contribution of the spatial and temporal components  of our proposed architecture, we present in \tabref{tab:abl} the results obtained when alternatively replacing the PSE by a CNN (CNN+TAE) or the TAE by a GRU (PSE+GRU).

\vspace{-.29cm}
\paragraph{Contribution of the PSE:}
As seen in \tabref{tab:abl}, the PSE accounts for an increase of $1.7$ points of mIoU compared to the CNN-based model (CNN+TAE). This supports both the hypothesis that CNNs are only partly relevant on medium-resolution images, and that considering the image as an unordered set of pixels is a valid alternative. Not only does this approach yield better classification performance, but it also circumvents the problem of image batching, which leads to faster data loading (see Table 2 in the appendix). Additionally, we train a TAE on pre-computed means and standard deviations of the spectral channels over the parcels (MS+TAE), which achieves a $48.9$ mIoU score. We can thus conclude that the PSE learns statistical descriptors of the acquisitions' spectra which are more meaningful than simple means and variances or convolutional features.
\vspace{-.29cm}
\paragraph{Design of the PSE:} 
We show in \tabref{tab:abl}, the performance of our architecture without geometric features $f$. The resulting $0.9$ point decrease in mIoU confirms that geometric information plays a role in the classification process. We note that, even without such features, our proposed approach outperforms the convolution-based model (CNN+TAE ).

We have tried replacing the handcrafted geometric features $f$ with a CNN operating on the binary mask of the parcel. However, the gains were minimal, and we removed this extra step for simplicity's sake.

Lastly, we tried training our architecture with a reduced number of sampled pixels ($S=16$, and $S=32$). The model maintains a good performance with an mIoU over $50$ points.  This indicates that the decrease in processing time and memory  could be further improved at the cost of a minor drop in precision.
\begin{table}[h]
\begin{center}
\begin{tabular}{l|cc}
                                                & O.A. & mIoU \\ \hline \toprule
PSE+TAE (ours)                     &\textbf{94.2} \footnotesize{\rpm 0.1}  &  \textbf{50.9}  \footnotesize{\rpm 0.8}  \\ \midrule
$\hat{q}=\upp{q}{T}$           & 94.2 \footnotesize{\rpm 0.1}    &   50.7  \footnotesize{\rpm 0.5}   \\
$S = 16$            & 94.3 \footnotesize{\rpm 0.2}    &   50.5  \footnotesize{\rpm 0.8}   \\
$\hat{q}=\max_t \upp{q}{t}$            & 94.2 \footnotesize{\rpm 0.2}    &   50.3  \footnotesize{\rpm 0.7}   \\
$S = 32$           & 94.2 \footnotesize{\rpm 0.1}    &   50.1  \footnotesize{\rpm 0.5}   \\

{No geometric features}            & 93.9 \footnotesize{\rpm 0.1}    &   50.0  \footnotesize{\rpm 0.7}   \\ \midrule

PSE+Transformer$+\hat{q}$                   &94.1 \footnotesize{\rpm 0.2}  &  49.5  \footnotesize{\rpm 0.7} \\
CNN+TAE                & 94.0 \footnotesize{\rpm 0.1}    &   49.2  \footnotesize{\rpm 1.1}   \\
MS+TAE                & 93.7 \footnotesize{\rpm 0.1}    &   48.9  \footnotesize{\rpm 0.9}   \\
PSE+GRU$+\textbf{p}$   & 93.6  \footnotesize{\rpm 0.2}  & 48.7 \footnotesize{\rpm 0.3}     \\

PSE+GRU     & 93.6  \footnotesize{\rpm 0.2}  & 47.3 \footnotesize{\rpm 0.3}     \\
PSE+Transformer                   &93.4 \footnotesize{\rpm 0.2}  &  46.6  \footnotesize{\rpm 0.9} \\
\end{tabular}
\end{center}
\caption{Ablation study of our different design choices, sorted by decreasing mIoU.}
\label{tab:abl}
\end{table}
\vspace{-.3cm}
\paragraph{Contribution of the TAE:}
Replacing the temporal attention encoder with a GRU (PSE+GRU) decreases the performance by $3.6$ points of mIoU (\tabref{tab:abl}). 
The TAE not only produces a better classification but also trains faster thanks to parallelization. 

Unlike the comparison between Transformer and RNNs architectures in \cite{russwurm2019self}, our modified self-attention mechanism 
extracts more expressive features than the RNN-based approach.

We also evaluate the influence of the positional encoding $p$ of the Transformer by adding $p$ to the input tensors of the GRU unit (PSE+GRU$+ \textbf{p}$). This reduces the gap with our method to $2.2$ points of mIoU. This shows that the improvement brought by the TAE is due to both its structure and the use of a positional encoding. 
\vspace{-.35cm}
\paragraph{Design of the TAE:}
In order to evaluate the benefits of our different contributions over the Transformer, we adapted the architecture presented in \cite{russwurm2019self} to use a PSE network instead of spectral means for embedding parcels (PSE+Transformer), for a performance $4.3$ points below our TAE.
By replacing the proposed temporal max-pooling by our our master query forming scheme (PSE+Transformer$+\hat{q}$), we observed an increase of $2.9$ points of mIoU. 
The remaining $1.4$ mIoU points between this implementation and ours can thus be attributed to our direct use of inputs to compute the TAE's output instead of a smaller intermediary value tensor.

Finally, we compare our mean-pooling strategy with max-pooling ($\hat{q}=\max_t \upp{q}{t}$) and computing the master query from the last element of the sequence ($\hat{q}=\upp{q}{T}$). While the mean query approach yields the best performance, the last element of the sequence in our dataset produces a meaningful query as well. However, this may not be the case for other regions or acquisition years.

\section*{Conclusion}
\vspace{-.1cm}
In this paper, we considered the problem of object-based classification from time series of satellite images. We proposed to view such images as unordered sets of pixels to reflect the typical coarseness of their spatial resolution, and introduced a fitting encoder.
To exploit the temporal dimension of such series, we adapted the Transformer architecture \cite{vaswani2017attention} for embedding time-sequences. We introduced a master query forming strategy, and exploited that our network learns end-to-end to simplify some operations.

Evaluated on our new open-access annotated benchmark of agricultural parcels, our method produces a better classification than all other re-implemented methods. Furthermore, our network is several times faster and more parsimonious in memory than other state-of-the-art methods such as convolutional-recurrent hybrid networks.
We hope that by mitigating some of the limitations of deep learning methods such as processing time and memory requirement, our approach would accelerate their adoption in real-life, large-scale Earth observation applications.

Our results suggest that attention-based models are an interesting venue to explore for analysing the temporal profiles of satellite time series, as well as other analogous vision tasks such as action recognition in videos. Likewise, set-based encoders are a promising and overlooked paradigm for working with the coarser resolutions of remote sensing applications.

\FloatBarrier 
{\small
\balance
\bibliographystyle{ieee_fullname}
\bibliography{psetae}
}

\ARXIV{
\clearpage
\nobalance
\appendix
\section*{Supplementary Material} 
\FloatBarrier 
\setcounter{figure}{0}
\setcounter{table}{0} 
We show the hyperparameters of the different competing methods in Table \ref{tab:hyperp_baselines}. We also provide a breakdown of the processing times during training for the different architectures in Table \ref{tab:perf_time}. The average time per  batch is decomposed into data loading time, forward pass and gradient back-propagation.

\begin{table}[h!]
\centering
\begin{tabular}{ll}
\multicolumn{2}{r}{Number of parameters} 
\\ \midrule
\textbf{CNN+GRU} &     $144 \, 204$            \\\toprule
\tabitem $3 \times 3$ convolutions: 32, 32, 64 kernels &  \\
\tabitem Global average pooling &  \\
\tabitem Fully connected layer: 128 neurons &  \\
\tabitem Hidden state size: 130  &  \\ \midrule
\textbf{CNN+TempCNN}       &        $156 \, 788$              \\ \toprule
\tabitem $3 \times 3$ convolutions: 32, 32, 64 kernels &  \\
\tabitem Global average pooling &  \\
\tabitem Fully connected layer: 64 neurons &  \\ 
\tabitem Temporal convolutions:  &  \\
32, 32, 64 kernels of size 3 &  \\
\tabitem Flatten layer \\ \midrule
\textbf{Transformer}       &        $178 \, 504$              \\ \toprule
\tabitem   $d_{k} = 32$, $d_{v} = 64$, $d_{model} = 128$, $d_{inner} = 256$\\
\tabitem $n_{head} = 4$, $n_{layer} = 1$\\ \midrule
\textbf{ConvLSTM}           &            $178 \, 356$     \\ \toprule
\tabitem Hidden feature maps: 64 & \\ \midrule
\textbf{RF} & \\ \toprule
\tabitem Number of trees: 100 & \\
\end{tabular}
\caption{Hyperparameters of the competing architectures. For all models we use the same values for the decoder $\MLP_{3}$.}
\label{tab:hyperp_baselines}
\end{table}

\begin{table}[!ht]
\centering
\begin{tabular}{l|c|ccc}
 Time in        &\multirow{2}{*}{Total} & \multirow{2}{*}{Loading} &  \multirow{2}{*}{Forward}  &  \multirow{2}{*}{Backward} \\
 ms/batch       &      & & \\\hline
PSE+TAE (ours)  & \bf 107  & \bf 85  &    11  & \textbf{11}\\
CNN+TempCNN  & 381 & 365   & \textbf{4} & 12  \\
CNN+GRU     & 437 & 365  &   14 & 58  \\
Transformer     & 8 & 1  &   2 & 5  \\
ConvLSTM    & 530  & 365 &  61 & 104\\
\end{tabular}
\caption{Comparison of processing time for different methods for batches of $128$ parcels. We can see that the processing time is dominated by the loading time except for the Transformer which processes pre-computed means.}
\label{tab:perf_time}
\end{table}
Lastly, for a more qualitative evaluation of our PSE+TAE architecture, we provide its confusion matrix on the test set on Figure \ref{fig:conf_mat} as well as a visual representation of its predictions on Figure \ref{fig:Prediction}.
\begin{figure}[h!]
    \centering
    \includegraphics[width=\linewidth]{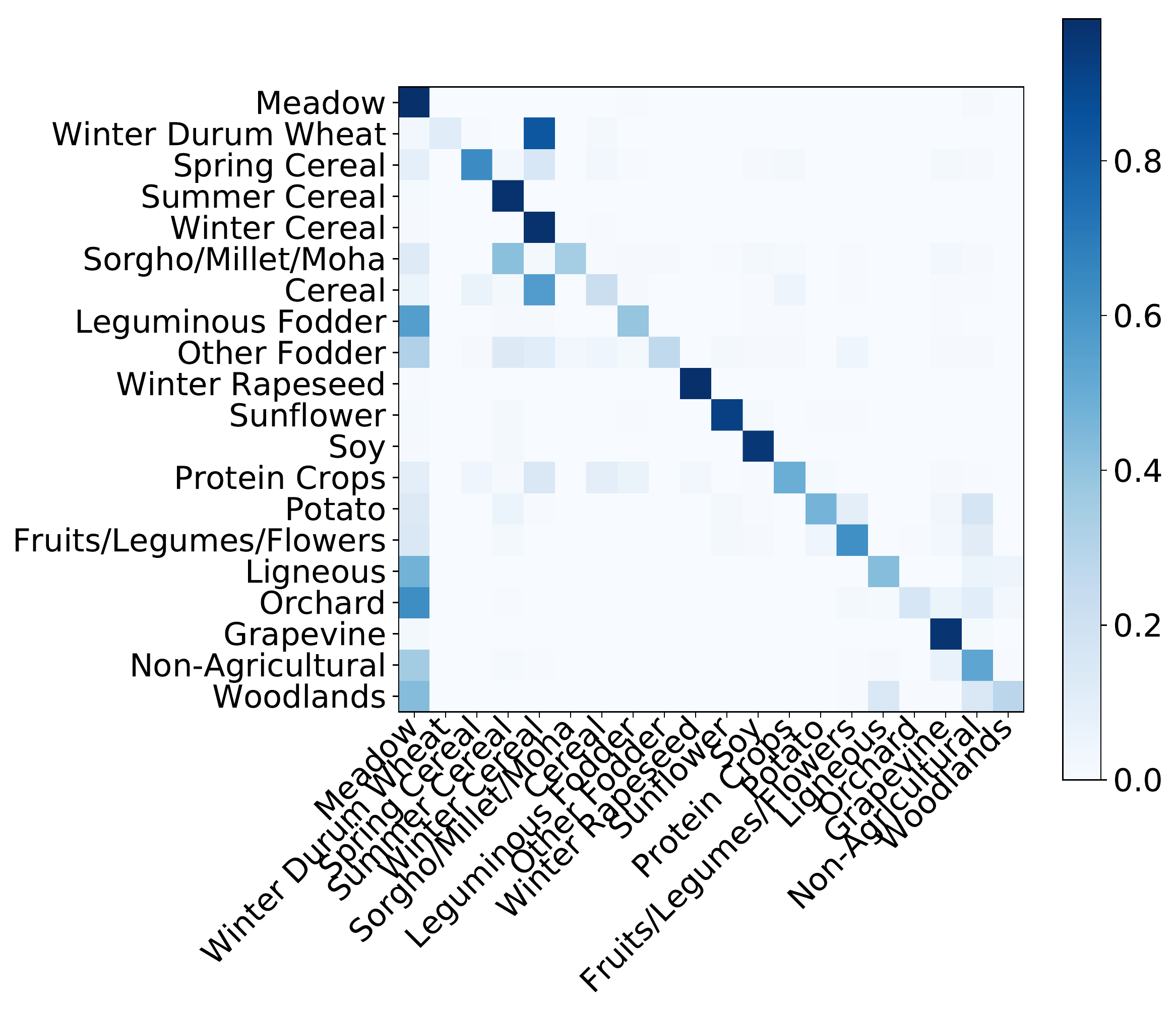}
    \caption{Confusion matrix for our PSE+TAE architecture on the AOI. The color represents the number of parcels, expressed relatively to the total population of the class they belong to. We note many of the errors are misclassification as \emph{Meadows}, the most represented class in our dataset. Additionally, the model struggles to discriminate between \emph{Winter Durum Wheat} and \emph{Winter Cereal}, likely due to their similar phenology.}
    \label{fig:conf_mat}
\end{figure}

\begin{figure*}
    \centering
    \includegraphics[trim=0cm 3cm 15cm 15cm]{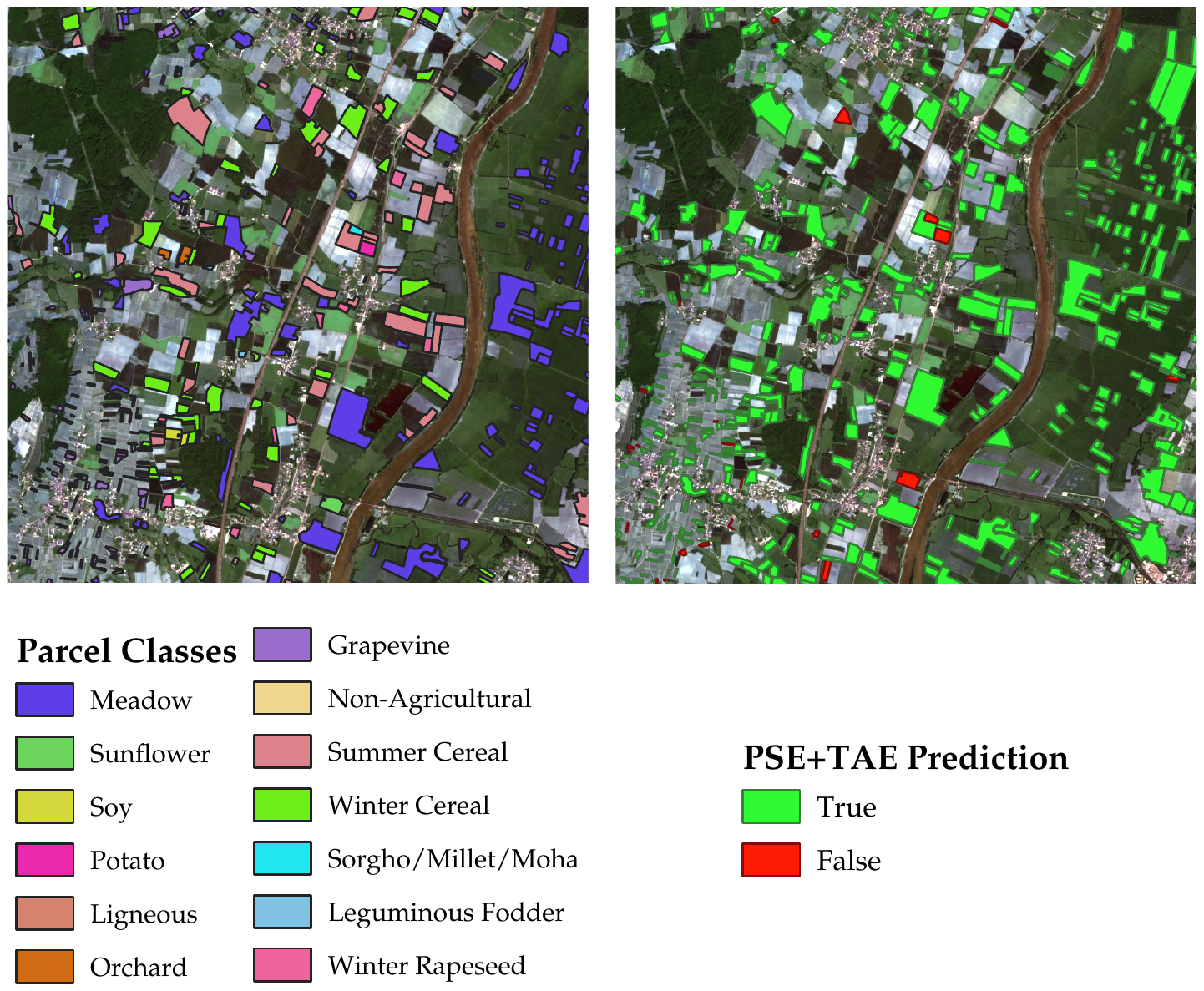}
    \caption{The left picture shows the true labels of a set of parcels, drawn from the test set. The right hand figure shows the parcels for which our PSE+TAE architecture produced a correct prediction in green and a false prediction in red. On both figures, the background corresponds to a Sentinel-2 observation (May 2017).}
    \label{fig:Prediction}
\end{figure*}

}{}

\end{document}